%% file: main.tex
\documentclass[twocolumn,pre,superscriptaddress]{revtex4}
\pdfoutput=1
\usepackage{amsmath,breqn,mathtools,amsfonts} 
\usepackage{hyperref}
\allowdisplaybreaks
\usepackage{graphicx}
\usepackage{tikz}
\usepackage{xcolor}

\DeclareMathOperator{\ExpOp}{\mathbb E}
\DeclarePairedDelimiterX{\ExpArg}[1]{[}{]}{#1}
\newcommand{\Exp}{\ExpOp\ExpArg*}

\makeatletter
\let\cat@comma@active\@empty
\makeatother

\begin{document}

\title{A Renormalization Group Approach to Connect Discrete- and Continuous-Time Descriptions of Gaussian Processes}
\author{Federica Ferretti}
\affiliation{Dipartimento di Fisica, Universit\a`{a} Sapienza, 00185 Rome, Italy}
\affiliation{Istituto Sistemi Complessi, Consiglio Nazionale delle Ricerche, UOS Sapienza, 00185 Rome, Italy}
\author{Victor Chard\`es}
\affiliation{Laboratoire de Physique de l'\a'{E}cole Normale sup\a'{e}rieure (PSL University), CNRS, Sorbonne Universit\a'{e}, Universit\a'{e} de Paris, 75005 Paris, France}
\author{Thierry Mora}
\affiliation{Laboratoire de Physique de l'\a'{E}cole Normale sup\a'{e}rieure (PSL University), CNRS, Sorbonne Universit\a'{e}, Universit\a'{e} de Paris, 75005 Paris, France}
\author{Aleksandra M Walczak}
\affiliation{Laboratoire de Physique de l'\a'{E}cole Normale sup\a'{e}rieure (PSL University), CNRS, Sorbonne Universit\a'{e}, Universit\a'{e} de Paris, 75005 Paris, France}
\author{Irene Giardina}
\affiliation{Dipartimento di Fisica, Universit\a`{a} Sapienza, 00185 Rome, Italy}
\affiliation{Istituto Sistemi Complessi, Consiglio Nazionale delle Ricerche, UOS Sapienza, 00185 Rome, Italy}
\affiliation{INFN, Unit\a`{a} di Roma 1, 00185 Rome, Italy}

\date{\today}

\begin{abstract}

\input{abstract}

\end{abstract}


\maketitle

\section{Introduction}
\input{intro}

\section{linear second order processes}
\input{arma21}

\subsection{Fixed points}
\input{fixed_points}

\section{Generalization to ARMA($p,q$) processes}\label{sec:general}
\input{general}


\section{Conclusion}
\input{conclusion}

\begin{acknowledgments}
We thank A. Vulpiani, M. Baldovin, A. Cavagna for helpful conversations, and A. C. Costa for useful comments and discussions. 
 This work was partially supported by the ERC Consolidator Grant n. 724208, ERC Advanced Grant n.785932 and by the Italian Ministry of Foreign Affairs and International Cooperation through the Adinmat project.
\end{acknowledgments}

\appendix

\section{}\label{app:A}
\input{fixed-points}

\section{}\label{app:Dpoint}
\input{Dpoint}

\section{}\label{app:B}
\input{appB}

\section{}\label{effar2}
\input{effective-ar2}

\bibliography{bibliography}

\clearpage
\onecolumngrid

\end{document}

%% file: abstract.tex
Discretization of continuous stochastic processes is needed to numerically simulate them or to infer models from experimental time series. However, depending on the nature of the process, the same discretization scheme, if not accurate enough, may perform very differently for the two tasks. Exact discretizations, which work equally well at any scale, are characterized by the property of invariance under coarse-graining. Motivated by this observation, we build an explicit Renormalization Group approach for Gaussian time series generated by auto-regressive models. 
We show that the RG fixed points correspond to discretizations of linear SDEs, and only come in the form of first order Markov processes or non-Markovian ones.
This fact provides an alternative explanation of why standard delay-vector embedding procedures fail 
in reconstructing partially observed noise-driven systems. We also suggest a possible effective Markovian discretization for the inference of partially observed underdamped equilibrium processes based on the exploitation of the Einstein relation.

%% file: intro.tex
How to properly represent continuous-time stochastic processes by discrete-time descriptions is a central problem in applied science. Models used in all areas of physics frequently take the form of stochastic differential equations (SDEs), yet the experimental observation of any real process produces discrete sequences of data points. Moreover, numerical integration of any SDE require to define transition probabilities over finite time steps. Hence finding a good discretization scheme is key for both numerical integration and parametric inference of continuous-time processes. 

Since exact solutions are not available for arbitrary processes, the general strategy is to resort to a Taylor-It\^o expansion of the integrated process. The lowest order expansion is the Euler-Maruyama scheme, which is (strongly) convergent as $\tau^{1/2}$, being $\tau$ the discretization time step  \cite{Platen-Kloeden}. 
This discretization scheme has been widely used in the literature, due to its simplicity and intuitive interpretation (it accounts to estimating velocities as finite differences). However, it has also been observed that the Euler-Maruyama approximation cannot be employed in the derivation of parametric inference methods for second or higher order processes, as it leads to the extraction of inconsistent parameter estimators \cite{Pedersen, prx, Lehle2018, Lehle2015,Gloter, Samson-EulerContrast, hypoellyptic-DitSamson, clairon-samson, pokern}. In other words, if we straightforwardly apply this scheme to learn the underlying continuous process from the discrete available data, we get the wrong result. On the contrary, adopting higher order discretization schemes, consistent inference algorithms (both Bayesian and non Bayesian) can be designed. 

The failure of the Euler-Maruyama discretization (\emph{Euler} for short) can be explained in the following terms: in order to learn second or higher order stochastic models, the dynamical information we need to extract from the fluctuations is of higher order (in $\tau$) than the accuracy with which those are reconstructed by the Euler scheme.
Nonetheless, the correlation functions that we reconstruct from a Euler simulation faithfully reproduce those of the integrated continuous-time process, at arbitrary order --- provided that the simulation time step ($\tau_{\rm sim}$) is sufficiently small, compared to the typical time scales of the system 
($\tau_{\rm obs}$). 
While this separation of scales is possible in numerical simulations, state space inference requires to discretize the process over the same time scales as those over which we observe the correlations.
The lack of such separation of scales explains the bad performance of the Euler discretization in parametric inference tasks, compared to numerical integration,
and suggests that as we `zoom out' our lens for the observation of the process, 
the dynamics recovers the correct statistical properties that were originally missing.

This property has motivated the introduction of data augmentation techniques in parametric inference, like \cite{Elerian, Eraker}, which consist of introducing and marginalizing additional intermediate states between pairs of observed points. It also reminds us of what happens in statistical field theory in the context of the Renormalization Group (RG). The bare theory, which provides an effective description of the system at the microscopic scale, might be incomplete; however, when we apply the RG procedure to get an effective theory at larger scales, this generates the missing terms and yields a model that correctly describes the large scale statistical fluctuations.
More precisely, the augmentation strategy of \cite{Elerian, Eraker} or the `zoom out' operation described above amount to temporal coarse-graining, which is reminiscent of the coarse-graining implemented in Real Space RG, with time steps playing the role of lattice spacings in the usual setting. 


In this paper, we identify the formal framework to develop this analogy with the RG approach, and we exploit it to recognize different classes of discrete-time stochastic processes based on their property of invariance under the RG transformation. The condition of invariance under RG amounts to requiring that, when the discretization interval $\tau$ is small enough, integrating the process 
over a single step $2\tau$ or through a combination of two steps on intervals $\tau$ should provide the same result.

We restrict our analysis to stationary Gaussian processes, for which an explicit RG map can be easily derived. In the space of these discrete-time models, Euler discretizations of linear SDEs are a subset. The study of the RG map reveals an interesting structure, where the fixed points point out classes of ``natural physical processes". They include full observations of continuous Markov processes, or partial observations of higher-dimensional processes inheriting a non-Markovian structure. No intermediate situation can be obtained. Higher-order Markov processes, whose transition probabilities depend on two or more previous observations, are not RG fixed points. This result underscores the lack of finite-dimensional delay vector embeddings for stochastic systems.


The paper is organized in the following way: in Sec. II we consider as a starting point the linear damped Langevin equation, and illustrate in this simple case how an RG procedure can be defined and applied to discretizations of the continuous process. We derive the RG map, identify the fixed points, and give their interpretation. We also discuss the possibility of building effective Markovian embeddings and their limitations, an issue relevant for inference purposes. In Sec. III we extend the previous analysis to arbitrary higher order linear processes, and deduce more general conclusions. Finally, in Sec. V we summarize and discuss our results.


%% file: arma21.tex
\subsection{A simple case}

The 
shortcomings of the Euler scheme mentioned in the Introduction are manifested for processes of second or higher order. By $n$-th order processes we mean that they are described by $n$-th order SDEs or, equivalently, that they are obtained from the partial observation of $n$-dimensional stochastic processes, whose structure is such that the noise is transferred from the hidden coordinates to the single observed degree of freedom (\emph{hypoelliptic} diffusions). Partial observations of such noise-driven systems break the Markovianity, introducing temporal noise correlations and memory effects in the description of the observed dynamics \cite{Zwanzig-book, Miguel1980}. These features are not captured by the Euler discretization scheme, which has the peculiar property of reducing to a Markovian discrete-time model when applied to this kind of systems. 

The simplest example, for $n=2$, is a linear damped Langevin equation in which only the positional coordinates are directly measured. Let us consider 
\begin{flalign}
&dx = vdt,\qquad \label{stoch-harm} \\
&dv = -\eta v dt - \kappa x dt + \sigma dW, \label{stoch-harm1}
\end{flalign}
with $W(t)$ a Wiener process. We assume that the coordinate $x$ is observed at a finite sampling rate $\tau^{-1}$, producing an infinite time series denoted as $\{X_n, n\in \mathbb N\}$. For this process, under stationary conditions, an exact solution can be computed and used in inference and simulation problems \cite{Kalman,Gillespie}; however, we are interested here in understanding how discrete models are related to their continuous counterparts. We therefore discretize Eqs.~\eqref{stoch-harm}--\eqref{stoch-harm1}, i.e. we integrate the continuous equations over a time interval of length $\tau$ and expand at the first order in $\tau$ the resulting integral expressions (Euler). 
Eliminating the $v$ variable, we obtain a discrete update equation for the $x$ coordinate that has the following structure:
\begin{equation}
X_{n} = \psi X_{n-1} + \theta X_{n-2} + \mu\epsilon_n,
\label{AR2}
\end{equation}
where $\psi=2-\eta\tau -\kappa\tau^2$, $\theta= (-1+\eta\tau)$, $\mu =\sigma\tau^{3/2}$ and $\epsilon_n \sim \mathcal N(0,1)$.

 Eq.~\eqref{AR2} is an autoregressive model of order two, denoted AR(2) \cite{Brockwell-Davis-book}, and it is fully characterized by the conditional probability $P(X_n|X_{n-1},X_{n-2})$. 
Thanks to the Markovian structure of the discrete process in Eq.~\eqref{AR2}, the probability of the associated time series reads: 
\begin{equation}
P(\{X_n,n\in \mathbb N\}) = \prod_{n\geq2}P(X_n|X_{n-1},X_{n-2})P(X_0,X_1).
\end{equation}
Under stationary assumptions, we can move the initial condition arbitrarily far in the past in order to neglect boundary terms. The resulting $P(\{X_n\})$ can be interpreted as the Boltzmann weight of a configuration of spins on an infinite one-dimensional lattice with first- and second-nearest-neighbor interactions. 
The analogy will be helpful to derive an explicit RG map.

\subsection{The RG construction}

The failure of the Euler discretization \eqref{AR2} in inference approaches 
signals that it misses some important information about fluctuations on the local scale $\tau$. The questions we address are the following: is there a way, starting from Eq.\eqref{AR2}, to understand what ingredients are missing? 
Are there specific constraints on the coefficients of discrete equations like \eqref{AR2}, to ensure compliance with any continuous model?
 
We start our analysis by noticing that a good discretization of the original continuous process must be so independently on the precise value of the time interval $\tau$ we consider (as long as it is small enough). If we use the same scheme to integrate over a single step $2 \tau$ or through a combination of two steps on intervals $\tau$, the result should be the same. 
As we iterate this argument many times, we compel the structure of the discrete equations to remain unaltered at different scales. We look indeed at the discrete process on scales that are larger and larger than the original one, until the ratio between the characteristic observation time scale and the discretization time step becomes infinite ($\tau_{\mathrm {obs}}/\tau_{\mathrm{sim}}\to \infty$). In this way, upon redefinition of the time units, we approach the continuum limit. This procedure allows us to check whether a given discrete scheme is scale invariant, and therefore faithfully describes a reference continuous equation, 
or --- in case invariance is 
violated --- how the approach to the continuum limit occurs.

\begin{figure}
\includegraphics[width=\columnwidth]{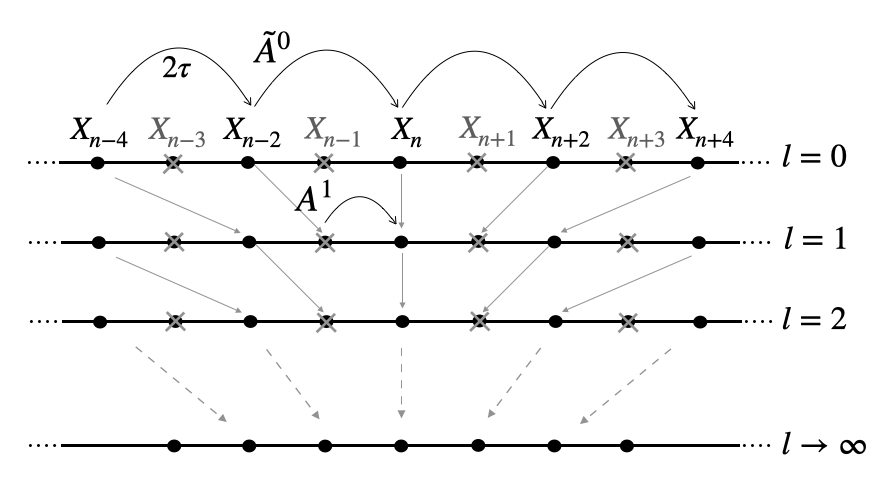}
\caption{Sketch of the RG transformation for a time series. The discretely observed coordinates $\{X_n\}$ play the role of spin variables on a lattice, with nearest neighbor and next-nearest neighbor coupling. The time step $\tau$ is the lattice spacing. Model parameters, $A^0$, are transformed into $\tilde A^0$ through coarse-graining, then rescaled to $A^1$ through the $\tau\to\tau/2$ operation.}
\label{fig:sketch}
\end{figure}

Let us formalize this idea using the language of the Renormalization Group. We explore the continuum limit through a progressive increase in the number of steps contained in a fixed time window. This is obtained by iterating the two operations that make up the RG: (i) coarse graining and (ii) joint rescaling of the time unit and of the parameters of the model. The procedure is sketched in Fig.~\ref{fig:sketch}.

We apply the method to Eq.\eqref{AR2}. Given the analogy of Euler time series with linear spin chains, we adopt the strategy of decimation to coarse grain \cite{Migdal-Kadanoff,Parisi-book,Brezin}: the goal is to get rid of half of the sites (e.g. odd ones) in the sequence generated by Eq.~\eqref{AR2}, thus deriving effective update equations for the even sub-series. 
In order to implement this transformation, we take a suitable linear combination of neighboring update equations of the form of \eqref{AR2}:
\begin{equation}
\text{Eqn}(X_{n})+\psi\text{Eqn}(X_{n-1})-\theta\text{Eqn}(X_{n-2}),
\label{lin-comb}
\end{equation}
which results into an update equation for 
the subseries:
\begin{equation}
X_{n} = \tilde\psi X_{n-2} + \tilde \theta X_{n-4} + \tilde r_n,
\label{tilde-eq}
\end{equation}
with $\tilde \psi = \psi^2+2\theta$, $\tilde\theta=-\psi^2$. The structure of Eq.\eqref{tilde-eq} looks similar to the one of Eq.\eqref{AR2} with updated parameters. There is, however, a crucial difference: unlike in the original process, the random increment 
\begin{equation}
	\tilde r_n=\mu\left[\epsilon_n +\psi\epsilon_{n-1}-\theta\epsilon_{n-2}\right]
\end{equation} 
is now correlated across nearest neighbors: $\Exp{\tilde r_n \tilde r_{n\pm2}} \neq 0$. This fact is better seen if we appropriately rewrite the random increment as a different linear combination of Gaussian variables: 
\begin{equation}
	\tilde r_n = \tilde\mu\tilde\epsilon_n + \tilde\nu\tilde\epsilon_{n-2},
\end{equation}
with $\tilde\epsilon_i\sim\mathcal N(0,1)$ new I.I.D. variables, and $\tilde\mu$, $\tilde\nu$ satisfying:
\begin{flalign}
	&\Exp{ \tilde r_n^2} = \tilde\mu^2 + \tilde\nu^2 = (1+\psi^2+\theta^2)\mu^2 ;\\
	&\Exp{ \tilde r_n \tilde r_{n\pm2}} = \tilde\mu \tilde\nu = -\theta\mu^2. 
	\label{mu-nu-def}
\end{flalign}
It can be verified that $\Exp{\tilde r_n^l \tilde r_{n\pm2k}^l}=0$ for $k>1$.  Thus Eq.~\eqref{tilde-eq} becomes: 
\begin{equation}
  X_{n} = \tilde\psi X_{n-2} + \tilde \theta X_{n-4} + \tilde \nu \tilde \epsilon_{n-2}+ \tilde \mu \tilde \epsilon_n,
  \label{tilde-eq2}
\end{equation}
also known as ARMA(2,1) model \cite{Brockwell-Davis-book}.
General autoregressive moving-average processes of order $(p,q)$, denoted ARMA($p,q$), are time series generated by update equations of the form:
\begin{equation}
X_n = \sum_{i=1}^p\phi_iX_{n-i} +\sum_{i=1}^q \nu_i \epsilon_{n-i} + \mu\epsilon_n,
\label{arma-pq-def}
\end{equation}
with $\nu_i, \mu \in \mathbb R$ and $\epsilon_n\sim \mathcal N(0,1)$ I.I.D.. The autoregressive (AR) part of the equation corresponds to the contribution from the previous $p$ states of the system; the moving average (MA) part, of order $q$, corresponds to the second sum in the RHS of Eq.~\eqref{arma-pq-def}, and is responsible for the time correlation of random increments. 

\begin{figure}
\includegraphics[width=\columnwidth]{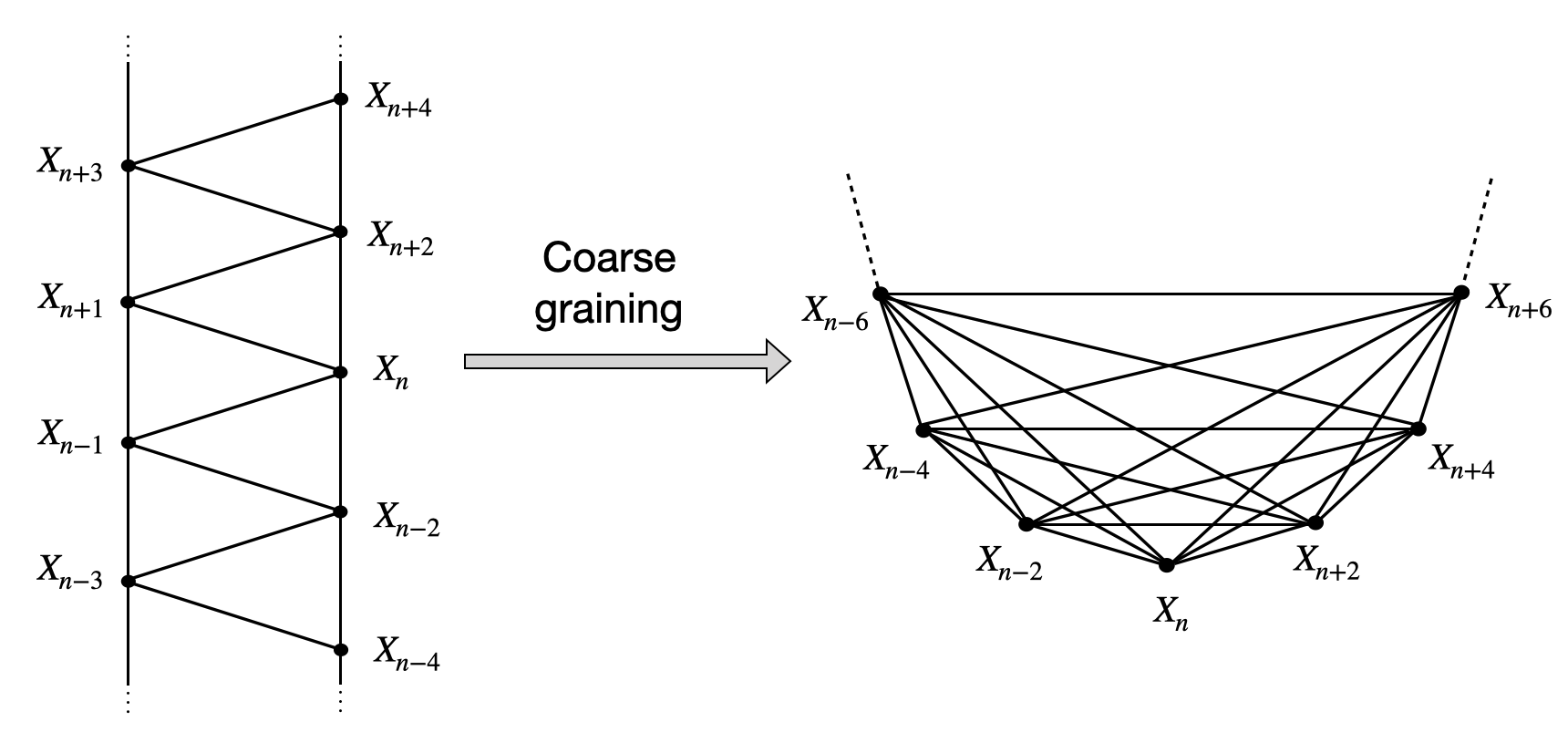}
\caption{Coarse-graining by decimation maps the ladder topology, corresponding to our AR(2) model, to a fully connected net. The new couplings are however not independent due to the non-trivial correlation properties of the random contributions.}
\label{fig:ladder}
\end{figure}

It is evident that, after coarse-graining, the process in Eq.~\eqref{tilde-eq} is no longer Markovian, as its probability distribution cannot be factorized into the product of single-step transition probabilities.
Recalling the analogy with the spin chain, RG creates, since the first iteration, infinite-range effective couplings, starting with just first- and second-nearest-neighbor bonds. This is indeed the effect of simple decimation on the zig-zag ladder topology, to which the AR(2)  process of Eq.~\eqref{AR2} corresponds (Fig.~\ref{fig:ladder}). However, these emerging couplings are not independent, and the four parameters in Eq.~\eqref{tilde-eq2} are sufficient to characterize them.

In order to get a closed-form RG transformation, we need to ensure that further iterations of the decimation procedure do not keep introducing novel higher order terms. Luckily, the ARMA(2,1) structure is stable, as an example of a more general result that we discuss in the following (Sec.~\ref{sec:general}). If we call  $\mathcal A=(\psi,\theta,\mu,\nu)$ the set of parameters of a generic ARMA(2,1) model, and $\tilde {\mathcal A}=(\tilde\psi, \tilde\theta, \tilde\mu, \tilde\nu)$ the parameters of the coarse grained model obtained after decimation, then applying RG yields a well-defined map from $\mathcal A$ to $\tilde{\mathcal A}$.

The second operation that completes the RG iteration is rescaling the time step, $2\tau\to\tau$, and reabsorbing this change of units through a redefinition of the parameters.
The parameters of our models are dimensionless, yet their dependency on $\tau$ is what determines how to connect any discrete process to its continuous-time counterpart.
We  express $\psi,\ \theta$ as asymptotic power series of $\tau$, $\psi(\tau)=\sum_k\psi_k\tau^k$ and $\theta(\tau)=\sum_k\theta_k\tau^k$, and work, up to the desired order, with recursive relations for the coefficients of the series expansion, $\psi_k$ and $\theta_k$. These coefficients are now dimensional and get rescaled with the time unit. The same idea can be applied to $\nu$ and $\mu$ by expanding them in powers of $\tau^{1/2}$. 
It is then convenient to reparametrize the noise amplitudes as:
\begin{equation}
\alpha \coloneqq \Exp{r_n r_n} = \mu^2+\nu^2;\quad
\beta \coloneqq \Exp{r_n r_{n\pm1}} = \mu\nu,
\label{ab-def}
\end{equation}
since their asymptotic series expansion involves  integer powers of $\tau$:
$\alpha(\tau) = \sum_k \alpha_k\tau^k$, $\beta(\tau) = \sum_k\beta_k\tau^k$.

The physical dimension of each coefficient $A_k \in \{\psi_k,\theta_k,\alpha_k,\beta_k\}$ is now set by the order of the corresponding term in the series expansion. Each of them gets rescaled, after coarse graining, as  $A_k^{l+1}=2^{-k}\tilde A_k^l$, where $l$ is an index counting the RG iterations.
This operation defines the RG map as a set of recursive equations in an infinite-dimensional parameter space, made up of the Taylor coefficients parametrizing the 4 functions $\psi^l(\tau),\, \theta^l(\tau),\, \alpha^l(\tau),\, \beta^l(\tau)$:
\begin{flalign}
 \label{rec-AR1}\psi_k^{l+1} &= 2^{-k}\left[2\theta_k^l + \sum_{i=0}^k\psi_i^l\psi_{k-i}^l\right],\\
 \label{rec-AR2}\theta_k^{l+1} &= -2^{-k} \sum_{i=0}^k\theta_i^l\theta_{k-i}^l.
\end{flalign}
Similar equations (reported in App.~\ref{app:A}) for the coefficients $\alpha_k$ and $\beta_k$ are obtained from the following definition of the coarse-grained parameters $\tilde\alpha^{l}(\tau)$ and $\tilde\beta^{l}(\tau)$:
\begin{flalign}
\label{tilde-a}&\tilde\alpha^{l} = \left[1+(\psi^{l})^2 +(\theta^{l})^2\right]\alpha^{l} + 2\psi^{l}(1-\theta^{l})\beta^{l};\\
\label{tilde-b}&\tilde\beta^{l} = \psi^{l}(1-\theta^{l})\beta^{l} - \theta^{l}\alpha^{l}.
\end{flalign}
Notice that since the recursion equations at order $k$ only involve lower orders, they can be solved recursively over $k$, and can also be truncated to an arbitrary order while retaining a closed form.

%% file: fixed_points.tex
\begin{table*}[t] 
\caption{\label{tab:fixed_points} Fixed point solutions of the RG recurrence relations up to third order in $\tau$. We find 4 manifolds of fixed points, corresponding to 4 types of processes, parametrized by the arbitrary constants $u,\, s,\, z$ and $b$.
In addition to the reported ones, there are diverging fixed points.} 
\begin{ruledtabular}
\begin{tabular}{l c | c c  c c  c c  c c | c c  c c  c c  c c}
\centering
 & & \multicolumn{8}{c|}{\small{\textbf{AR coefficients}}} & \multicolumn{8}{c}{\small{\textbf{MA coefficients}}}\\
&Model & $\psi_0^*$ & $\theta_0^*$ & $\psi_1^*$ & $\theta_1^*$ & $\psi_2^*$ & $\theta_2^*$ & $\psi_3^*$ & $\theta_3^*$ & $\alpha_0^*$ & $\beta_0^*$ & $\alpha_1^*$ & $\beta_1^*$ & $\alpha_2^*$ & $\beta_2^*$ & $\alpha_3^*$ & $\beta_3^*$ \\
\hline
A.&MA(0) & 0 & 0 & 0 & 0 & 0 & 0 & 0 & 0 & $s$ & 0 & 0 & 0 & 0 & 0 & 0 & 0\\
B.&AR(1) & 1 & 0 & $u$ & 0 & $u^2/2$ & 0 & $u^3/6$ & 0 & 0 & 0 & $s$ & 0 & $us$ & 0 & $2u^2s/3$ & 0 \\
C.&ARMA(2,1) & -1 & -1 & $u$ & $2u$ & $-u^2/2$ & $-2u^2$ & $u^3/6$ & $(2u)^3/6$ & 0 & 0 & $4s$ & $s$ & $-8us$ & $-2us$ & $32u^2s/3$ & $13u^2s/6$\\
D.&ARMA(2,1) & 2 & -1 & $u$ & $-u$ & $z$ & $-u^2/2$ & $u(6z-u^2)/12$ & $-u^3/6$ & 0 & 0 & $-2s$ & $s$ & $-2us$ & $us$ & $4b - (2z+3u^2)s$ & $b$\\
\end{tabular}
\end{ruledtabular}
\end{table*}

\begin{figure}
\includegraphics[width=\columnwidth]{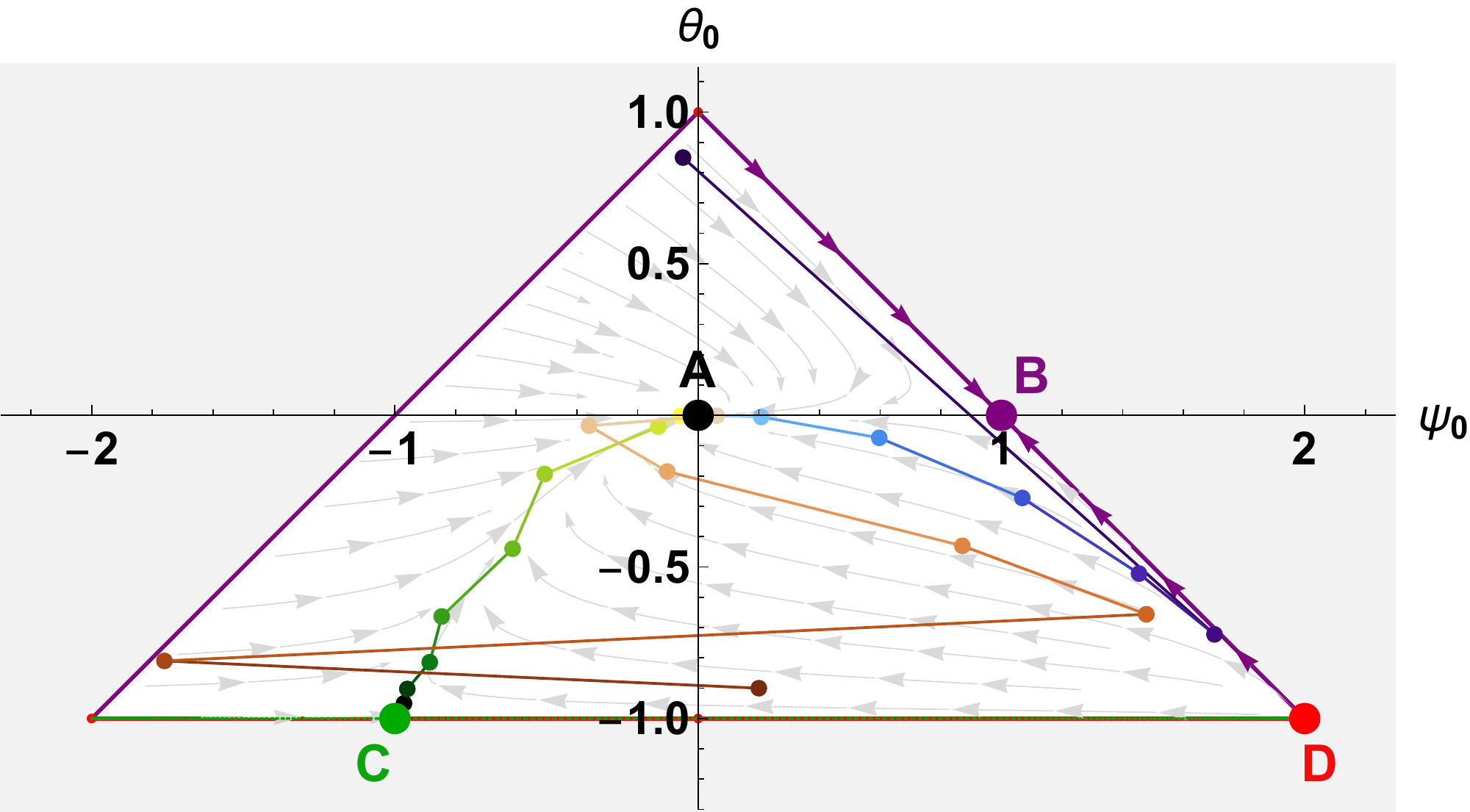}
\caption{Fixed point manifolds of the RG map, projected on the plane of leading autoregressive coefficients $\psi_0,\, \theta_0$. 
The interior of the triangle is the basin of attraction of fixed point A. The equal sides are the basin of attraction of point B. The basins of points C and D are contained in the basis of the triangle (plus vertex on the top for D).
The Euler AR(2) process has coordinates (2,-1) in this plane. Color-shaded trajectories represent the solution of recurrence relations from three sample initial conditions. Arrows show the direction of (discrete) moves and should not be interpreted as continuous flow lines.} 
\label{fig:fixed_points}
\end{figure}
Our interest is in finding the fixed points of the recursion relations defined above. RG fixed points capture indeed the scale-invariant 
behavior of system. Table~\ref{tab:fixed_points} reports the solutions for the parameters $\mathcal A^*$ of the ARMA(2,1) model, showing their coefficients $A^*_k$ up to order $k=3$.
There are four manifolds of fixed points in this space, corresponding to different classes of stochastic processes. Their projection 
onto the plane of leading order AR coefficients, $\psi_0$ and $\theta_0$, is shown in Fig.~\ref{fig:fixed_points}. 

The first class of fixed points A corresponds to sequences of independent random variables ($\psi=\theta=\nu=0$). 
Fixed points B are AR(1) processes, i.e. processes where only two time points are involved and there is no memory in the noise term: they can be interpreted as discretizations of linear first-order SDEs of the form:
\begin{equation}
dx = uxdt+\sqrt{s}dW.
\label{1ord-fp}
\end{equation}
By induction on the fixed point equations for the $A_k$ coefficients ($k\in \mathbb N$), it is possible to prove that the fixed point parameters of the B process are: $\theta^*(\tau)=\beta^*(\tau)=0$, $\psi^*(\tau)=e^{u\tau}$, $\alpha^*(\tau)=\frac{s}{2}(e^{2u\tau}-1)$.

In addition, there are two other fixed-point manifolds corresponding to ARMA(2,1) models, denoted by C and D. C is not a continuous process, but evolves through finite jumps.  The AR coefficients, which appear in the deterministic part of the equation, read $\psi(\tau)=-e^{-u\tau}$ and $\theta(\tau)=-e^{-2u\tau}$. The process has a three-branched phase diagram, and evolves in time jumping from one branch to the next one -- approaching the origin or moving away from it depending on the sign of $u$. This structure is invariant under the RG transformation we defined, but it is precisely determined by the details of the decimation procedure. If the coarse graining was implemented differently, for instance by trimming two points out of three, the same type of process would not be a fixed point. 

Finally, the D class represents the discretization of a partially observed two-dimensional process of the form:
\begin{flalign} 
\label{x-cont-2}dx &= vdt +\sigma_xdW_x,\\
\label{v-cont-2}dv &= -\eta v dt -\kappa x dt +\sigma dW_v,
\end{flalign}
of which \eqref{stoch-harm} is a particular case \footnote{In fact the fixed point D includes a more general class of two-dimensional continuous-time processes of the form 
$d \bold y = \bold A\bold y + \bold Bd\bold W$, as described in App.~\ref{app:Dpoint}. 
All these processes are however projected, after marginalization of the velocities, to the same ARMA(2,1) model. What changes is just the mapping from the coefficients of the continuous-time process $\bold A$ and $\bold B$ to those of the ARMA(2,1) model (clearly noninvertible).}.
The $(\psi,\theta)$ coefficients of D 
reconstruct a second derivative at leading order, and the effect of linear drift at first order (with $u=-\eta$ in Table \ref{tab:fixed_points}). The variable $s=-\sigma_x^2$ in Table \ref{tab:fixed_points} encodes noise added to the $x$ variable,
and is zero in inertial models like \eqref{stoch-harm}. 

Inertial models are of special interest. In this case, since $s=0$, noise contributions are determined by third-order coefficients. 
If we start with the Euler model \eqref{AR2} as an initial condition for the RG recurrence relations, whose noise coefffcients are $\alpha_k^0=\sigma^2\delta_{k,3},\beta_k^0= 0$, the associated RG flow, on the plane of MA parameters, reads:
\begin{equation}
	\alpha_3^{l} = \sigma^2\left(\frac{2}{3} + \frac{1}{3}4^{-l}\right);\quad \beta_3^{l} = \sigma^2\left(\frac{1}{6}-\frac{1}{6}4^{-l}\right).
\label{sol-ab}
\end{equation}
Upon renormalization, the Euler time series \eqref{AR2} falls into class D, with $b=\sigma^2/6$ and $s=0$. 
While the initial model is strongly convergent as $\tau^{1/2}$, the resulting fixed point is at least convergent as $\tau^{3/2}$. The asymptotic values $(\alpha_3^{\infty},\beta_3^{\infty})$ are those we would obtain if we applied a higher order discretization scheme like \cite{LI} to \eqref{stoch-harm} in the first place. A more exhaustive analysis of fixed point D and of its parameters can be found in Appendix \ref{app:Dpoint}.

To summarize the analysis performed so far, we have shown that the Euler discretization of the inertial equation we considered is {\it not} invariant upon RG transformations. This encapsulates the fact that Euler does not appropriately describe  the statistical properties of the continuous model on the original discretization scale. We also showed that, as RG progresses, Euler leads to effective discrete models that are, on the contrary, invariant. This convergence occurs exponentially fast 
--- see Eq.~\eqref{sol-ab} --- and explains why numerical simulations performed with the Euler scheme work. 
Finally we note that higher order discretizations of Eqs. \eqref{stoch-harm}--\eqref{stoch-harm1} already possess an ARMA(2,1) structure and are therefore invariant upon RG transformations. This explains why they provide good algorithms for local (in time) inference approaches, in contrast to Euler.

\subsection{Effective Markov description} 
Although we discussed that Euler-based maximum likelihood inference approaches for stochastic dynamical systems are inconsistent, numerical and analytical evidence \cite{prx, Pedersen, Lehle2018, Lehle2015, GloterOU} hints at a possible \emph{effective} Markov discretization of the form of \eqref{AR2} for generalized Langevin equations at equilibrium. Such discretization neglects noise correlations, but employs a rescaled damping coefficient $\eta'=(2/3)\eta$. In other words, the information missed by the Euler scheme at the discretization scale $\tau$ can be effectively reabsorbed into a single rescaling factor.

For linear processes, this behavior can be interpreted as following from the Einstein relation, which imposes a well-known relationship between the drag coefficient and the noise amplitude (through temperature and mass) \cite{Kubo,Vulpiani-response} --- see App. \ref{effar2}. 
In our case, this relation involves the parameters $\psi_1$ and $\alpha$. 
If we select the fixed-point value of the diagonal entry of the noise covariance matrix, $\alpha^*_3=(2/3)\sigma^2$, instead of the Euler value $\alpha^0_3=\sigma^2$, and neglect off-diagonal elements, then we find by imposing the Einstein relation that we must also rescale $\eta$ by $2/3$. 
In Appendix \ref{effar2} we show how this effective Euler-like discretization can be derived 
in the simple case of a linear Langevin equation like Eq. \eqref{stoch-harm}.
Of course, this effective Markov description can only be used for maximum likelihood inference at the scale of $\tau$. The effective discretization gives, by construction, consistent maximum likelihood estimators, but cannot be iterated in simulations to reproduce the process on longer scales. Because of the convergence of the Euler-Maruyama integration algorithm, the simulated process would be the underdamped Langevin model with a damping coefficient equal to $\eta'=(2/3)\eta$.

In the context of continuous processes, the idea of building an effective Markov model that reproduces the statistics of a non-Markov linear model (only at the stationary level) has been very recently investigated in \cite{Broedersz-Markov}.

%% file: general.tex
The RG procedure we detailed for ARMA(2,1) models can be generalized to arbitrary ARMA$(p,q)$ processes, defined by Eq.~\eqref{arma-pq-def}. Decimation of the time series can be done by combining $p$ neighboring equations of the form of \eqref{arma-pq-def}, in a similar way as in Eq.~\eqref{lin-comb} (see App.~\ref{app:B}).
This decimation step yields a new update equation with an ARMA($p$,$\tilde q$) structure, where
\begin{equation}
\tilde q=\left\lfloor\frac{p+q}{2}\right\rfloor,
\label{q'}
\end{equation}
and $\lfloor \cdot \rfloor$ denotes the rounding down operation.
The condition of invariance under RG imposes that fixed points satisfy $\tilde q=q$, 
implying $q = p$ or $q = p - 1$.

This result leads to two important observations. First, it shows that purely autoregressive models of order $p\geq 2$, AR($p$)=ARMA($p$,0) (i.e. models with no memory in the noise), cannot be stable points and thus cannot be exact discretizations of stochastic differential equations of second or higher order. As a consequence of this fact is that for partially observed processes ($p\geq2$) a Markov description of the dynamics ($q=0$) is impossible. Second, the sharp selection of $q$ reveals that longer memory than $p$ is irrelevant in the RG sense.

Our finding is related to the non-existence of exact delay vector embeddings for noise-driven systems. 
Embedding approaches consist of stacking a finite number of subsequent points to define a new dynamical variable $\mathcal{X}=(X_{n-p+1},\ldots,X_n)$ --- known as delay vector \cite{Takens,Casdagli} --- and assuming that it follows a Markov dynamics. An equivalent approach is to estimate the derivatives of the observed stochastic process through differences of subsequent measurements: the Euler discretization is a simple example of this procedure. 
In both cases, the Markov dynamics of the embedded process is described by AR($p$) models. In conclusion, although the delay-vector method is standard in deterministic contexts, it cannot be directly extended to stochastic processes, as partial observation sets strong limitations to phase space reconstruction for stochastic dynamical systems \cite{Stark1,Stark2,OnsagerMachlup53,BorraBaldovin-chaos,ACC}.

%% file: conclusion.tex
In this work we investigated the connection between discrete and continuous descriptions of stochastic Gaussian processes. Using an RG approach, we studied how the properties of given discretization schemes change as we change the observation scale of the process, from the original discretization step, up to to the continuum limit $\tau/\tau_{obs}\to 0$. We started our analysis focusing on the simple case of a linear damped Langevin equation, and then generalized the procedure to higher order processes. In this broader framework, we defined an RG map on the space of ARMA($p,q$) models, i.e. a class of generative models for Gaussian processes, and identified exact discretizations of continuous-time stochastic processes through its fixed points. 
Our results underscore the impossibility to describe partially observed dynamics through an effective Markov process, but also show that longer memory than the order of the process ($q>p$) is irrelevant.

The issue relates to the embedding problem and is especially relevant for the design of parametric inference methods. A possible suggestion that can be drawn from our analysis is to abandon the Markov setting in favor of descriptions with correlated noise, by introducing an additional noise delay for any new coordinate in the delay vector. The RG construction shows indeed that time correlations in the noise terms are spontaneously generated to match the original dimension of the partially observed dynamical system.
Consistently, inference algorithms that take into account the right noise correlations since the beginning correctly estimate the parameters of the underlying continuous model, without the need of augmentation techniques. \cite{Pedersen, prx, Lehle2018, Lehle2015,Gloter}.

Nonetheless, we also showed that, up to a parameter rescaling, effective discretizations based on second-order Markov models --- AR(2) --- can still be used for the inference of damped equilibrium processes. 
This observation raises the question of whether alternative (e.g. variational) RG transformations exist that can map AR(2) models into other AR(2) processes, allowing for an effective memoryless description of Gaussian processes. How to extend these results to nonlinear processes also remains an open question.

%% file: fixed-points.tex
In this Appendix we look in greater details at the fixed points of the recursive equations. Our starting point are Eqs. \eqref{rec-AR1}--\eqref{tilde-b}. Since the recursive equations for $\psi,\theta$ are independent of those for $\alpha$ and $\beta$, it is convenient to start with the coefficients of the autoregressive (AR) part of the model, and then focus on the stochastic (MA) contribution at a second stage.

\subsection*{AR coefficients}
From Eqs.~\eqref{rec-AR1}--\eqref{rec-AR2}, at each order $k$ of the series expansion, we obtain a two-dimensional map from which the fixed points for the considered coefficients, $(\psi_k^*,\theta_k^*)$, can be extracted. Notice that higher order recurrence relations are only coupled to lower order ones, so one can solve them iteratively. Here we are just interested in the study of the fixed points.

The recursive relations at the leading order ($k=0$) read:
\begin{flalign}
\label{map0-psi}\psi_0^{l+1} &= (\psi_0^{l})^2 + 2\theta_0^{l}\\
\label{map0-theta}\theta_0^{l+1} &= -(\theta_0^{l})^2.
\end{flalign}
As it is evident from Eqs.~\eqref{rec-AR1}--\eqref{rec-AR2}, only at this order the system is nonlinear; at any subsequent order the map is linear. There are 4 fixed points for the nonlinear recurrence relations~\eqref{map0-psi}--\eqref{map0-theta}:
\begin{enumerate}
\item[A.] $(\psi_0^*,\theta_0^*) = (0,0)$. This is a process with null autoregressive part. 
Moving average contributions are specified by the recurrence relations for the MA coefficients. 
\item[B.] $(\psi_0^*,\theta_0^*) = (1,0)$.  This fixed point corresponds to a first order process of the form $X_n  = X_{n-1} + r_n$, i.e. an ARMA(1,$q$) process; the moving average order $q$ will be determined by the recurrence relations for $\alpha_k$ and $\beta_k$.
\item[C.] $(\psi_0^*,\theta_0^*) = (-1,-1)$. This is an ARMA(2,$q$) process, of the form $X_n = -X_{n-1}-X_{n-2}+r_n$. Here again, the random contribution will be specified by the recursion relations for $\alpha_k$ and $\beta_k$. 
\item[D.] $(\psi_0^*,\theta_0^*) = (2,-1)$. This model is a specific case of an ARMA(2,$q$) model, known as ARIMA(1,1,$q$) model \cite{Brockwell-Davis-book}. 
Thanks to the specific value assumed by the coefficients of the AR part, one can indeed rewrite the process as $(1-L)^2X_{n} = r_n$, where $L$ is the lag operator: $LX_n = X_{n-1}$. So $(1-L)$ is the discrete differencing operator and $\{(1-L)X_n\}=\{\bar V_n\}$ is the reconstructed velocity series. We leave once again the value of $q$ unspecified for the moment, as it is determined by the analysis of the recurrence relations for the MA coefficients.
\end{enumerate}
Notice that only the fixed point A is an asymptotically stable point, whose basin of attraction is the interior of the triangle in Fig.~\ref{fig:fixed_points}. The other points are unstable, at least in some directions. 

We are especially interested in the class of discrete-time models represented by point D, since, at the leading order, the model can be considered as a discretized version of a second order SDE of the kind $\ddot x = \xi$, with $\xi(t)$ a Gaussian noise (white or colored, depending on the MA coefficients). Higher order contributions can modify the coefficients $\psi$ and $\theta$ in front of $X_{n-1}$ and $X_{n-2}$, but they will not affect the interpretation of such a process as a discretization of a second order SDE. The only difference will be in the addition of position- or velocity-dependent linear forces.

Proceeding to the next order, $k=1$, the study of the fixed points reveals that their number and their nature (i.e. the kind of dynamical models they correspond to) is left unchanged. The AR coefficients of the A fixed point are still constrained to zero ($\psi_1=\theta_1=0$) while a new free parameter ($u$ in Table~\ref{tab:fixed_points}) appears at order $k=1$ for the other fixed points, in order to accomodate the arbitrariness of what can be interpreted as a linear force in the overdamped case B, or a damping force in case D.

Let us now examine what happens at order $k=2$.
%
%
Given the recurrence relations:
\begin{flalign}
\psi_2^{l+1} &= \frac{1}{4}\left[2\psi_0^{l}\psi_2^{l} + (\psi_1^{l})^2 + 2\theta_2^{l}\right],\\
\theta_2^{l+1} &= -\frac{1}{4}\left[2\theta_0^{l}\theta_2^{l}+(\theta_1^{l})^2\right],
\end{flalign}
the fixed points, at this order, are expanded as follows:
\begin{enumerate}
\item[A.] $(\psi_0^*,\theta_0^*,\psi_1^*, \theta_1^*,\psi_2^*,\theta_2^*)=(0,0,0,0,0,0).$ 
\item[B.] $(\psi_0^*,\theta_0^*,\psi_1^*,\theta_1^*,\psi_2^*,\theta_2^*)=(1,0,u,0,u^2/2,0) $. 
\item[C.] $(\psi_0^*,\theta_0^*,\psi_1^*,\theta_1^*,\psi_2^*,\theta_2^*)=(-1,-1,u,2u,-u^2/2,-2u^2) $. 
\item[D.] $(\psi_0^*,\theta_0^*,\psi_1^*,\theta_1^*,\psi_2^*,\theta_2^*)=(2,-1,u,-u,z,-u^2/2)$. Compared to the previous fixed points, we have here an additional arbitrariness on $\psi_2^*$. The novel free parameter $z$ accounts for an $x$-dependent force in the second-order continuous-time SDE, of which the fixed point model D can be considered a discretization. Indeed the discretization picture above keeps holding, even if the ARIMA(1,1,$q$) structure is lost in favor of an ARMA(2,$q$) one, when $\psi_2^*\neq u^2/2$, in general.
For points B and C, on the contrary, no new free parameter appears and the values of $\psi_2$ and $\theta_2$ are fixed by the values of lower order coefficients. In these cases, the dimension of the fixed point manifold does not change by moving from $k=1$ to $k=2$.
\end{enumerate}

Of course the study of the fixed points can be developed to any desired order, but the discussion we made so far is already sufficient to characterize their autoregressive nature. Indeed, it can be proven by induction that the AR order of the fixed points is left unchanged at subsequent orders in the expansion.

\subsection*{MA coefficients}

Recurrence relations for the MA coefficients $\mu$ and $\nu$ or, equivalently, for the parameters $\alpha$ and $\beta$, depend on those of the AR coefficients as shown above. The great advantage of working with $\alpha$ and $\beta$ is that, when the AR coefficients $\psi$ and $\theta$ are fixed, their recurrence relations are linear: 

\begin{widetext}
\begin{flalign}
\tilde \alpha_k^{l} &= (1+\psi_0^2+\theta_0^2)\alpha_k^l + 2 \beta_k^l\psi_0(1-\theta_0) + \sum_{i=0}^{k-1} \left[ \alpha_i^l \sum_{j=0}^{k-i} \left(\psi_j\psi_{k-i-j}+\theta_j \theta_{k-i-j}\right)  + 2\beta_i^l\psi_{k-i}(1-\theta_0) - 2\beta_i^l\sum_{j=0}^{k-i-1}\psi_j\theta_{k-i-j}\right]\label{MA1}\\
\tilde \beta_k^{l} &= \beta_k^l\psi_0(1-\theta_0)-\alpha_k^l\theta_0-\sum_{i=0}^{k-1}\alpha_i^l \theta_{k-i}+\sum_{i=0}^{k-1}\beta_i^l \left[\psi_{k-i}(1-\theta_0)-\sum_{j=0}^{k-i-1}\psi_j\theta_{k-i-j}\right].\label{MA2}
\end{flalign}
\end{widetext}
We analyze their behavior for each of the 4 fixed point classes we found above.

At the leading order we have a homogeneous system:
\begin{flalign}
\alpha_0^{l+1} &= \left(1+\psi_0^2+\theta_0^2\right)\alpha_0^{l} + 2\psi_0(1-\theta_0)\beta_0^{l}, \\
\beta_0^{l+1} &= \psi_0(1-\theta_0)\beta_0^{l} - \theta_0\alpha_0^{l}.
\end{flalign}
The single fixed point of such system is the origin $(\alpha_0^*,\beta_0^*)=(0,0)$, unless the coefficients $\theta_0$ and $\phi_0$ take values that render the fixed point equations linearly dependent. This happens only for class A, where the condition $(\psi_0^*,\theta_0^*)=(0,0)$ leaves $\alpha_0^*$ as a free parameter, while $\beta_0^*=0$. As a result, $q=0$, and this fixed point model is just a sequence of I.I.D. Gaussian variables. 
For  autoregressive models (i.e. fixed points B, C and D), the condition $(\alpha_0^*,\beta_0^*)=(0,0)$ tells us that $\tau$-independent noise contributions are prohibited. This feature enables us to interpret these models as discretizations of SDEs 
\cite{Gardiner}. 

Moving to a first order expansion (the lowest nontrivial one for the last three fixed points), the recursion relations \eqref{MA1}--\eqref{MA2} take the form of a 2D affine map.
For the four fixed points we find:
\begin{enumerate}
\item[A.] MA($q$): $(\alpha_0^*,\beta_0^*,\alpha_1^*,\beta_1^*)=(s,0,0,0)$. 
\item[B.] ARMA(1,$q$): $(\alpha_0^*,\beta_0^*,\alpha_1^*,\beta_1^*)=(0,0,s,0)$, with $s$ a real parameter. As expected, even at this order $q=0$ and the process reduces to a simple AR(1) model.
\item[C.] ARMA(2,$q$): $(\alpha_0^*,\beta_0^*,\alpha_1^*,\beta_1^*)=(0,0,4s,s)$, $s\in \mathbb R$. We have a manifold of fixed points, represented by a line on the plane of first order covariance coefficients.
\item[D.] AR(I)MA(2,$q$): $(\alpha_0^*,\beta_0^*,\alpha_1^*,\beta_1^*)=(0,0,-2s,s)$. Again, we have a line of fixed point solutions. The corresponding process can be interpreted as the first order discretization of a partially observed SDE, as we discuss more in details in Appendix \ref{app:Dpoint}.
\end{enumerate}

At second order, the structure of the MA part of the fixed points remains almost the same: A and B remain memoryless processes ($q=0$), and no new parameters appear for C and D. Therefore we report directly the result for $k=3$, the last order where arbitrariness can be introduced to modify the structure of the fixed points:
\begin{enumerate}
\item[A.] MA(0): $(\alpha_0^*,\beta_0^*,\alpha_1^*,\beta_1^*,\alpha_2^*,\beta_2^*,\alpha_3^*,\beta_3^*) = (s,0,0,0,0,0,0,0) $. 
\item[B.] ARMA(1,0): $(\alpha_0^*,\beta_0^*,\alpha_1^*,\beta_1^*,\alpha_2^*,\beta_2^*,\alpha_3^*,\beta_3^*) = (0,0,s,0,u s,0,\frac{2}{3}\psi_1^2s,0) $. 
\item[C.] ARMA(2,1): $(\alpha_0^*,\beta_0^*,\alpha_1^*,\beta_1^*,\alpha_2^*,\beta_2^*,\alpha_3^*,\beta_3^*) = (0,0,4s,s,-8us,-2us,\frac{32}{3}u^2s,\frac{13}{6}u^2s) $. 
\item[D.] AR(I)MA(2,1): $(\alpha_0^*,\beta_0^*,\alpha_1^*,\beta_1^*,\alpha_2^*,\beta_2^*,\alpha_3^*,\beta_3^*) = (0,0,-2s,s,-2us,us,4b-(2z+3u^2)s,b) $. For this nontrivial fixed point, the recurrence relations at this order are linearly dependent, and admit infinitely many solutions. We parametrize them by taking $\beta_3^*=b$. 

\end{enumerate}

%% file: Dpoint.tex
This section is devoted to a more detailed discussion of the fourth fixed point, which, due to its physical meaning, we think deserves a special focus. 
We have already highlighted that, at leading order, the AR coefficients reproduce a second time derivative through the double differencing operator $\Delta^2=(1-L)^2$, with $L$ the lag operator: $LX_n = X_{n-1}$. This fact gives to the model an `integrated process' nature at leading order, also known as ARIMA(1,1,1) \cite{Brockwell-Davis-book}. Since the relation 
\begin{equation}
	\lim_{\tau\to0}\frac{\psi-\psi_0^*}{\tau} = -\lim_{\tau\to0}\frac{\theta-\theta_0^*}{\tau}
\end{equation}
also holds, this integrated process structure is kept up to $O(\tau)$. Deviations from it occur at higher order and are due to the presence of linearly $x$-dependent forces. 

The fixed point model is indeed the consistent discretization of a general class of partially observed continuous-time processes of the form:
\begin{equation}
d\bold y = \bold A \bold y dt + \bold B d\bold W,
\label{general-2}
\end{equation}
where $\bold y = (x,v)^{\top}$, $\bold W = (W_x,W_v)^{\top}$ with $W_x(t)$ and $W_v(t)$ independent Wiener processes and
\begin{equation}
\bold A = \begin{pmatrix}
-\lambda & 1 \\
-\kappa & -\eta
\end{pmatrix}, \quad \bold B \bold B^{\top}= \begin{pmatrix}
\sigma_{xx}^2 & \sigma_{xv}^2 \\
\sigma_{vx}^2 & \sigma_{vv}^2 
\end{pmatrix}. 
\label{param}
\end{equation}
Notice that setting $A_{12}=1$ does not imply a loss of generality: compared to $A_{12}=a\in \mathbb R\setminus\{0\}$, it just corresponds to a rescaling of the time unit, which does not alter the process (the only caveat is that $a<0$ would revert the time direction). The case $A_{12}=0$ is not of interest for us, as it would decouple the dynamics of the unobserved degrees of freedom from that of the observed ones. The entries in the drift matrix $\bold A$ must satisfy the stability condition, i.e., assuming that time evolves in the positive direction, $(t\geq0)$, it must be negative semidefinite. Finally, we have $\sigma_{xv}^2=\sigma_{vx}^2$ for the symmetry of the diffusion matrix.

Partial observation of the process in Eq.~\eqref{general-2} yields a Gaussian process described, at the continuous level, by a Generalized Langevin Equation (GLE), and, at the discrete level, by an ARMA(2,1) model. The GLE can be obtained by integrating the continuous equation for the $v$ variable and plugging it into the first one in \eqref{general-2}:
\begin{widetext}
\begin{equation}
dx(t) = \left\{e^{-\eta t} v(0) - \kappa \int_0^t dt' e^{-\eta(t-t')}x(t') + \int_0^t \left[B_{21}dW_x(t') + B_{22}dW_v(t')\right]\right\}dt - \lambda x(t) + B_{11}dW_x(t)+ B_{12}dW_v(t).
\label{GLE-gen}
\end{equation}
\end{widetext}
This equation contains a parametric dependence on the initial condition $v(0)$, which can be removed during discretization as in \cite{LI, prx}. 
Alternatively, an exact discrete update equation in $x$-space can be found from the exact integration of Eq.\eqref{general-2}. It reads:
\begin{widetext}
\begin{dmath}
X_n - \frac{\left(e^{2\bold A\tau}\right)_{12}}{\left(e^{\bold A\tau}\right)_{12}}X_{n-1} - \left[ \left(e^{2\bold A\tau}\right)_{11} - \frac{\left(e^{2\bold A\tau}\right)_{12}}{\left(e^{\bold A\tau}\right)_{12}}\left(e^{\bold A\tau}\right)_{11} \right] X_{n-2}  =  r_n,
\label{exact-discr}
\end{dmath}
with 
\begin{dmath}
r_n = \int_{t_n-2\tau}^{t_n} \left(e^{\bold A(2\tau-s)}\right)_{11}\left[B_{11}dW_x(s) + B_{12}dW_v(s)\right] - \frac{\left(e^{2\bold A\tau}\right)_{12}}{\left(e^{\bold A\tau}\right)_{12}} \int_{t_n-2\tau}^{t_n-\tau}  \left(e^{\bold A(\tau-s)}\right)_{11}\left[B_{11}dW_x(s) + B_{12}dW_v(s)\right] + 
\int_{t_n-2\tau}^{t_n} \left(e^{\bold A(2\tau-s)}\right)_{12}\left[B_{21}dW_x(s) + B_{22}dW_v(s)\right] - \frac{\left(e^{2\bold A\tau}\right)_{12}}{\left(e^{\bold A\tau}\right)_{12}} \int_{t_n-2\tau}^{t_n-\tau}  \left(e^{\bold A(\tau-s)}\right)_{12}\left[B_{21}dW_x(s) + B_{22}dW_v(s)\right].
\label{rn}
\end{dmath}
\end{widetext}
Eq.~\eqref{exact-discr} corresponds to an ARMA(2,1) process, since Eq.\eqref{rn} implies that $\Exp{r_nr_m}=\alpha \delta_{n,m} + \beta \delta_{n,m\pm1}$. 

It is possible to explicitly work out the calculation to find the 
mapping from $(\bold A,\bold B\bold B^{\top})$ to $(\psi,\theta,\alpha,\beta)$ --- and hence find the relation between Eq.~\eqref{param} and the variables parametrizing the fixed point D in Table \ref{tab:fixed_points}.
Let us start by performing a small $\tau$ expansion for the AR coefficients:
%
\begin{widetext}
\begin{flalign}
\label{psi_map} \psi &= \frac{\left(e^{2\bold A\tau}\right)_{12}}{\left(e^{\bold A\tau}\right)_{12}} 
\simeq 2- (\eta+\lambda)\tau + \frac{1}{2}\tau^2\left(-2\kappa+\eta^2+\lambda^2\right) + \frac{1}{6}\tau^3\left(3\kappa(\eta+\lambda)-\eta^3-\lambda^3\right);
\\
\label{theta_map} \theta &= \left(e^{2\bold A\tau}\right)_{11} - \frac{\left(e^{2\bold A\tau}\right)_{12}}{\left(e^{\bold A\tau}\right)_{12}}\left(e^{\bold A\tau}\right)_{11} = -e^{-(\eta+\lambda)\tau}\simeq
-1 + (\eta+\lambda)\tau - \frac{1}{2}\tau^2(\eta+\lambda)^2 + \frac{1}{6}\tau^3(\eta+\lambda)^3.
\end{flalign}
\end{widetext}
Identifying the coefficients  $\psi_0\dots\psi_3$ and $\theta_0\dots\theta_3$, we deduce that the model parameters in Eq.~\eqref{param} and the fixed point parameters of process D in Table~\ref{tab:fixed_points} are linked by the following relation:
\begin{equation}
u = -(\lambda+\eta); \quad z = -\kappa + \frac{1}{2}(\eta^2+\lambda^2).
\label{map-AR}
\end{equation}
The algebra for the derivation of $\alpha$ and $\beta$ is more laborious but not complicated. The results are:
\begin{dmath}
\alpha \simeq 2\sigma_{xx}^2\tau - 2\sigma_{xx}^2(\eta+\lambda)\tau^2 + \frac{2}{3}\left[\sigma_{vv}^2+2\eta\sigma_{xv}^2+\sigma_{xx}^2(-\kappa+3\eta^2+3\eta\lambda+2\lambda^2)\right]\tau^3;
\end{dmath}
\begin{dmath}
\beta \simeq -\sigma_{xx}^2\tau + \sigma_{xx}^2\tau^2 + \frac{1}{6}\left[\sigma_{vv}^2 + 2\eta\sigma_{xv}^2 + \sigma_{xx}^2(2\kappa-3\eta^2-6\eta\lambda-4\lambda^2)\right]\tau^3.
\end{dmath}
Hence we can deduce:
\begin{flalign}
s &=-\sigma_{xx}^2; \\ 
b &= \frac{1}{6}\left[\sigma_{vv}^2 + 2\eta\sigma_{xv}^2 + \sigma_{xx}^2(2\kappa-3\eta^2-6\eta\lambda-4\lambda^2)\right].
\label{map-MA}
\end{flalign}


Notice that the mapping given by Eqs.~\eqref{map-AR} and \eqref{map-MA} is noninvertible. Because of the partial nature of the observation, multiple models are mapped to the same ARMA process. In other words, there is no bijection between the continuous-time Markovian description of the dynamical system and its experimental non-Markovian observation. A unique reconstruction of a set of first order SDEs is impossible. Extracting the parameters of an underlying continuous-time Markov model from time series analysis is therefore an arbitrary task, which postulates the choice of suitable hidden variables. 

A subclass of models contained in fixed point D is given by inertial processes, which take the form:
\begin{flalign}
dx &= vdt;\\
dv &= -\eta v dt - \kappa x dt +\sigma_{vv}dW_v.
\label{inertial-ex}
\end{flalign}
They are obtained by setting $\lambda=\sigma_{xv}^2=0$ 
and $\sigma_{xx}^2=0$. This condition implies that there are no $O(\tau^{1/2})$ stochastic contributions to the observed process: $s=0$ in Table \ref{tab:fixed_points}. 
Because of the absence of 
these contributions, applying a Euler discretization to \eqref{inertial-ex} gives rise to an AR(2) model. 

We can take this inconsistent discretization of the 2nd order SDE as an initial condition for the RG recurrence relations of MA coefficients.
Since $s=0$, we have null $\alpha_k$ and $\beta_k$ up to $k=3$. Third order recurrence relations are in this case:
\begin{equation}
\alpha_3^{l+1} = \frac{1}{8}\left[6\alpha_3^{l}+8\beta_3^{l}\right];\quad
\beta_3^{l+1} = \frac{1}{8}\left[\alpha_3^{l}+4\beta_3^{l}\right].
\label{inertial-rec-3}
\end{equation}
Solutions lye on parallel lines
$\alpha_3+2\beta_3=c$, with $c$ a constant fixed by the initial condition $(\alpha_3^{0},\beta_3^{0})=(\sigma_{vv}^2,0)$. The intersection with the fixed point line, $\alpha_3^*=4\beta_3^*$, identifies in our parameter space the model which is reached by repeatedly coarse-graining the starting discrete-time model. 

Thanks to linearity, one can also compute how the asymptotic point is approached. The solution of Eq.~\eqref{inertial-rec-3} is
\begin{flalign}
\alpha_3^{l} &= 4^{-l}\frac{1}{3}\left[\alpha_3^{0}-\beta_3^{0}\right]+\frac{2}{3}\left[\alpha_3^{0}+2\beta_3^{0}\right],\\
\beta_3^{l} &= 4^{-l}\left[\frac{2}{3}\beta_3^{0}-\frac{1}{6}\alpha_3^{0}\right]+\frac{1}{6}\left[\alpha_3^{0}+2\beta_3^{0}\right];
\label{sol-ab-3}
\end{flalign}
so the discrete model converges in an exponentially fast way to a consistent scheme which is strongly convergent as $O(\tau^{3/2})$.
There is then an `asymptotic upgrade' of the order of convergence of the scheme, at least in the linear case. 

%% file: appB.tex
In this Appendix we show how the memory selection rule $q = p$ or $q = p - 1$ emerges from the condition of invariance under RG of general ARMA($p,q$) processes.
Given the generative model
\begin{equation}
X_n = \sum_{i=1}^p \phi_i X_{n-i} + \sum_{i=1}^q \nu_i \epsilon_{n-i} + \mu\epsilon_n,
\label{arma-pq}
\end{equation}
with $\epsilon_n\sim\mathcal{N}(0,1)$, decimation of the time series is performed through the linear combinations:
\begin{equation}
\text{Eqn}(X_n) + \sum_{i=1}^p(-1)^{i+1}\phi_i \text{Eqn}(X_{n-i}), 
\end{equation}
which generalizes Eq.~\eqref{lin-comb}. Notice that this combination only depends on the AR order $p$. The resulting discrete-time model reads:
\begin{widetext}
\begin{dmath}
X_n = \sum_{i=1}^p\left[1+(-1)^i\right]\phi_iX_{n-i} + \sum_{i=1}^p(-1)^{i+1}\phi_i\sum_{k=1}^p\phi_kX_{n-k-i} + \underbrace{\mu_0\epsilon_n+ \sum_{j=1}^q\mu_j\epsilon_{n-j}+\sum_{i=1}^p(-1)^{i+1}\phi_i\left[\mu_0\epsilon_{n-i}+\sum_{k=1}^q\mu_k\epsilon_{n-i-k}\right]}_{\tilde r_n}.
\label{cg-pq}
\end{dmath}
The second sum in Eq.~\eqref{cg-pq} can be rewritten, with a rearrangement of terms, as 
\begin{equation}
\sum_{i=1}^p(-1)^{i+1}\phi_i\sum_{k=1}^p\phi_kX_{n-k-i} = \sum_{i=1}^p(-1)^{i+1}\phi_i\sum_{k=1}^p\phi_kX_{n-k-i} = \sum_{i=1}^p(-1)^{i+1}\left(\phi_i^2X_{n-2i} + 2\sum_{k=i}^{\lfloor\frac{p-1}{2}\rfloor}\phi_i\phi_{2k+1}X_{n-2k-2}\right).
\label{cast-final}
\end{equation}
\end{widetext}
Eq.~\eqref{cast-final} shows that, after the decimation, one maintains an autoregressive part of order $p$. The picture is not modified by 
the first sum, which only contributes up to an AR order $\lfloor p/2\rfloor$. 

Thus Eq.~\eqref{cg-pq} can be rewritten as:
\begin{widetext}
\begin{dmath}
X_n = \sum_{i=1}^{\lfloor p/2\rfloor}2\phi_{2i}X_{n-2i} + 
\sum_{i=1}^p(-1)^{i+1}\left(\phi_i^2X_{n-2i} + 2\sum_{k=i}^{\lfloor \frac{p-1}{2}\rfloor}\phi_i\phi_{2k+1} X_{n-2k-2}\right)+ \tilde r_n,
\label{cg-double}
\end{dmath}
\end{widetext}
where it is possible to recognize a structure of the following kind
\begin{equation}
X_n = \sum_{i=1}^p \tilde \phi_i X_{n-2i} + \tilde r_n.
\end{equation}
The process corresponds to an ARMA model with the same autoregressive order as the original one ($p$), but now each jump covers a time interval of doubled amplitude. 

Further manipulation of the sums in Eq.~\eqref{cg-pq} allows us to find formal expressions for the AR coefficients of the coarse grained process, $\tilde \phi_{i=1\dots p}$: 
\begin{widetext}
\begin{equation}
\begin{cases}
\tilde \phi_1 = 2\phi_2 +\phi_1^2 ;\\
\tilde \phi_i = 2\phi_{2i} + (-1)^{i+1}\phi_i^2 + 2\phi_{2i-1}\sum_{k=1}^{i-1}(-1)^{k+1}\phi_k \quad \text{ for}\quad 2\leq i \leq \left\lfloor\frac{p}{2}\right\rfloor-1;\\
\tilde \phi_i = (-1)^{i+1}\phi_i^2 \hspace{5.45cm} \text{ for}\quad \left\lfloor\frac{p}{2}\right\rfloor+1 \leq i \leq p.
\end{cases}
\end{equation}
The coarse-grained coefficient $\tilde \phi_{\lfloor p /2 \rfloor}$ takes a different form depending on $p$ being even or odd:
\begin{flalign}
p \text{ odd}: \qquad \tilde \phi_{\lfloor p/2\rfloor} &= 2\phi_{2\lfloor p/2\rfloor} + (-1)^{1+\lfloor p/2\rfloor}\phi_{\lfloor p/2\rfloor}^2 ;\\
p \text{ even}: \qquad \tilde \phi_{\lfloor p/2\rfloor} &= 2\phi_{2\lfloor p/2\rfloor} + (-1)^{1+\lfloor p/2\rfloor}\phi_{\lfloor p/2\rfloor}^2 + 2\phi_{2\lfloor p/2\rfloor-1}\sum_{k=1}^{\left\lfloor \frac{p-1}{2}\right\rfloor}(-1)^{k+1}\phi_k.
\end{flalign}
\end{widetext}

Let us now restart from Eq.\eqref{cg-pq} and focus on the random term $r_n$. Since linear combinations of Gaussian variables are still Gaussian, one can properly redefine the $\epsilon_m$'s and rearrange the coefficients in front of them to rewrite $\tilde r_n=\sum_{i=0}^{\tilde q}\tilde\mu_i\epsilon_{n-2i}$, where 
\begin{equation}
\tilde q=\left\lfloor \frac{p+q}{2}\right\rfloor.
\end{equation}

We deduce there are only 2 invariant scenarios for ARMA($p,q$) processes: $q=p$ or $q=p-1$. This fact tells us that partial (discrete) observation of continuous-time processes let memory emerge: each hidden degree of freedom increases by one the order of both the AR part and the MA part of the discrete model, thus introducing color.

%% file: effective-ar2.tex
%
%
In this appendix we motivate the exploitation of the seemingly universal $2/3$ rescaling factor in effective Markov models discussed in the main text,
and provide a physical interpretation for it, working out the reference problem of an integrated Ornstein-Uhlenbeck (OU) process. 

The integrated OU process is the simplest example of 2nd-order SDE for which the Euler-related inconsistency appears. It is described by Eq.~\eqref{stoch-harm} with $\kappa=0$, where we suppose we can only observe (with infinite accuracy) the inertial degree of freedom, at a sampling rate $\tau^{-1}$. Let us recall the notation for the time series of empirical observations $\{X_n\}_{n\in \mathbb N}$, and for the time series of reconstructed velocities $\{\bar V_n\}_{n\in \mathbb N}$, where 
\begin{equation}
\bar V_n = \frac{\left(X_{n+1}-X_n\right)}{\tau}.
\end{equation}
Let us also introduce the series of \emph{real} velocities $\{V_n\}_{n\in \mathbb N}$, corresponding to the one we would obtain if we were able to measure directly the velocity degree of freedom. Because the evolution of the $v$ variable is described by an independent 1st-order SDE when $\kappa=0$, the time series $\{V_n\}_{n\in \mathbb N}$ is described by an AR(1) process. On the contrary, the evolution of the $x$ variables is non Markovian and expressed, at the continuous level, via a generalized Langevin equation. Consequently, the time series $\{\bar V_n\}_{n\in \mathbb N}$ inherits a nonzero MA order, ending in an ARMA(1,1). 

Nonetheless, we may ask whether it is possible to describe it with an \emph{effective} AR(1) process, which would correspond to an effective AR(2) process for the $\{X_n\}_{n\in \mathbb N}$ series.
Let us write a putative AR(1) model for the series of reconstructed velocities:
\begin{equation}
	\bar V_n - (1-\alpha) \bar V_{n-1} = \sigma \epsilon_n, 
	\label{AR1-effective}
\end{equation}
where $\epsilon_n\sim\mathcal N(0,1)$ I.I.D. and $\alpha$, $\sigma$ are parameters to fix.  The goal is to find an effective memoryless discrete model for $\{\bar V_n\}_{n\in \mathbb N}$ that reproduces correctly the sufficient statistics used by Bayesian and non-Bayesian parametric inference approaches. The common characteristic of these approaches is that of being derived from a Taylor-It\^o expansion in the small $\tau$ limit. They just exploit local dynamical information to learn the laws governing the evolution of the system, typically carried by the first few elements of the autocovariance of the time series. 


In the case of AR(1) models, a sufficient statistics corresponds to the set $\mathcal S_1 = \left\{\Exp {\bar V_n^2},\Exp {\bar V_n \bar V_{n+1}}\right\}$, i.e. the autocorrelation functions at equal time and at a distance of one time step \cite{LeePresse}.  
We impose on the observables $\mathcal S_1$ the two following consistency conditions:
\begin{enumerate}
\item[i.] The stationary distribution of $\bar V_n$ is the Maxwell Boltzmann distribution at temperature $T$ ($k_B=1$):
\begin{equation}
	\Exp{\bar V_n^2} = T.
	\label{equip}
\end{equation}
\item[ii.] The relation between the reconstructed acceleration $\bar A_n = (\bar V_{n+1}-\bar V_n)/\tau$ and the reconstructed velocity $\bar V_n$ is the one we can exactly compute for the integrated Ornstein-Uhlenbeck process: 
\begin{equation}
\Exp{\bar A_n\vert \bar V_n} \mathop{\simeq}_{\tau\to 0} -\frac{2}{3}\eta \bar V_n.
\label{ped-relation}
\end{equation}
A detailed derivation is in \cite{Pedersen}. From this relation we immediately derive the condition to impose on the observables of interest:
\begin{equation}
	\Exp{\bar V_{n+1}\bar V_n} = \left(1-\frac{2}{3}\eta\tau\right) \Exp{\bar V_n^2}.
	\label{ped}
\end{equation}
\end{enumerate}
The self-correlation function $\Exp{\bar V_n \bar V_{n+k}}$ of an AR(1) process of the form of Eq.~\eqref{AR1-effective} is explicitly known:
\begin{equation}
	\Exp{\bar V_n \bar V_{n+k}} = \frac{(1-\alpha)^{|k|}\sigma^2}{1-(1-\alpha)^2}.
	\label{self-corr}
\end{equation}
Taking its value at $k=1$ and using Eq.~\eqref{equip} in Eq.~\eqref{ped} yields the expected result $\alpha=(2/3) \eta\tau$. Computing the self-correlation of the reconstructed velocities at $k=0$ and imposing equipartition from Eq.~\eqref{equip}, we set the value of $\sigma$:
\begin{equation}
	\sigma^2 = T \left[1-(1-\alpha)^2\right] \mathop{\simeq}_{\tau\to0}  2T\alpha.
	\label{Einstein}
\end{equation}
This is the celebrated Einstein relation.

In conclusion, it is possible to describe the sequence of measurements of an integrated OU process as an effective AR(2) series with a rescaled damping coefficient $\eta' = (2/3)\eta$. This result is intuitive and could have been grossly derived by imposing the Einstein relation (which comes from Eq.~\eqref{equip} alone), and selecting only the main diagonal of the covariance matrix of random increments to cancel color, thus implying $\sigma^2 = 2/3 (2T \eta\tau)$. 

%% file: main.bbl
\begin{thebibliography}{34}
\expandafter\ifx\csname natexlab\endcsname\relax\def\natexlab#1{#1}\fi
\expandafter\ifx\csname bibnamefont\endcsname\relax
  \def\bibnamefont#1{#1}\fi
\expandafter\ifx\csname bibfnamefont\endcsname\relax
  \def\bibfnamefont#1{#1}\fi
\expandafter\ifx\csname citenamefont\endcsname\relax
  \def\citenamefont#1{#1}\fi
\expandafter\ifx\csname url\endcsname\relax
  \def\url#1{\texttt{#1}}\fi
\expandafter\ifx\csname urlprefix\endcsname\relax\def\urlprefix{URL }\fi
\providecommand{\bibinfo}[2]{#2}
\providecommand{\eprint}[2][]{\url{#2}}

\bibitem[{\citenamefont{Platen and Kloeden}(1992)}]{Platen-Kloeden}
\bibinfo{author}{\bibfnamefont{P.~E. K.~E.} \bibnamefont{Platen}}
  \bibnamefont{and} \bibinfo{author}{\bibfnamefont{P.~E.}
  \bibnamefont{Kloeden}}, \emph{\bibinfo{title}{Numerical Solution of
  Stochastic Differential Equations}}, vol.~\bibinfo{volume}{23} of
  \emph{\bibinfo{series}{Stochastic Modelling and Applied Probability}}
  (\bibinfo{publisher}{Springer-Verlag Berlin Heidelberg},
  \bibinfo{year}{1992}), \bibinfo{edition}{1st} ed.

\bibitem[{\citenamefont{Pedersen et~al.}(2016)\citenamefont{Pedersen, Li,
  Gr\ifmmode~\u{a}\else \u{a}\fi{}dinaru, Austin, Cox, and
  Flyvbjerg}}]{Pedersen}
\bibinfo{author}{\bibfnamefont{J.~N.} \bibnamefont{Pedersen}},
  \bibinfo{author}{\bibfnamefont{L.}~\bibnamefont{Li}},
  \bibinfo{author}{\bibfnamefont{C.}~\bibnamefont{Gr\ifmmode~\u{a}\else
  \u{a}\fi{}dinaru}}, \bibinfo{author}{\bibfnamefont{R.~H.}
  \bibnamefont{Austin}}, \bibinfo{author}{\bibfnamefont{E.~C.}
  \bibnamefont{Cox}}, \bibnamefont{and}
  \bibinfo{author}{\bibfnamefont{H.}~\bibnamefont{Flyvbjerg}},
  \bibinfo{journal}{Phys. Rev. E} \textbf{\bibinfo{volume}{94}},
  \bibinfo{pages}{062401} (\bibinfo{year}{2016}),
  \urlprefix\url{https://link.aps.org/doi/10.1103/PhysRevE.94.062401}.

\bibitem[{\citenamefont{Ferretti et~al.}(2020)\citenamefont{Ferretti,
  Chard\`es, Mora, Walczak, and Giardina}}]{prx}
\bibinfo{author}{\bibfnamefont{F.}~\bibnamefont{Ferretti}},
  \bibinfo{author}{\bibfnamefont{V.}~\bibnamefont{Chard\`es}},
  \bibinfo{author}{\bibfnamefont{T.}~\bibnamefont{Mora}},
  \bibinfo{author}{\bibfnamefont{A.~M.} \bibnamefont{Walczak}},
  \bibnamefont{and} \bibinfo{author}{\bibfnamefont{I.}~\bibnamefont{Giardina}},
  \bibinfo{journal}{Phys. Rev. X} \textbf{\bibinfo{volume}{10}},
  \bibinfo{pages}{031018} (\bibinfo{year}{2020}),
  \urlprefix\url{https://link.aps.org/doi/10.1103/PhysRevX.10.031018}.

\bibitem[{\citenamefont{Lehle and Peinke}(2018)}]{Lehle2018}
\bibinfo{author}{\bibfnamefont{B.}~\bibnamefont{Lehle}} \bibnamefont{and}
  \bibinfo{author}{\bibfnamefont{J.}~\bibnamefont{Peinke}},
  \bibinfo{journal}{Phys. Rev. E} \textbf{\bibinfo{volume}{97}},
  \bibinfo{pages}{012113} (\bibinfo{year}{2018}),
  \urlprefix\url{https://link.aps.org/doi/10.1103/PhysRevE.97.012113}.

\bibitem[{\citenamefont{Lehle and Peinke}(2015)}]{Lehle2015}
\bibinfo{author}{\bibfnamefont{B.}~\bibnamefont{Lehle}} \bibnamefont{and}
  \bibinfo{author}{\bibfnamefont{J.}~\bibnamefont{Peinke}},
  \bibinfo{journal}{Phys. Rev. E} \textbf{\bibinfo{volume}{91}},
  \bibinfo{pages}{062113} (\bibinfo{year}{2015}),
  \urlprefix\url{https://link.aps.org/doi/10.1103/PhysRevE.91.062113}.

\bibitem[{\citenamefont{Gloter}(2006)}]{Gloter}
\bibinfo{author}{\bibfnamefont{A.}~\bibnamefont{Gloter}},
  \bibinfo{journal}{Scandinavian Journal of Statistics}
  \textbf{\bibinfo{volume}{33}}, \bibinfo{pages}{83} (\bibinfo{year}{2006}),
  \eprint{https://onlinelibrary.wiley.com/doi/pdf/10.1111/j.1467-9469.2006.00465.x},
  \urlprefix\url{https://onlinelibrary.wiley.com/doi/abs/10.1111/j.1467-9469.2006.00465.x}.

\bibitem[{\citenamefont{Samson and Thieullen}(2012)}]{Samson-EulerContrast}
\bibinfo{author}{\bibfnamefont{A.}~\bibnamefont{Samson}} \bibnamefont{and}
  \bibinfo{author}{\bibfnamefont{M.}~\bibnamefont{Thieullen}},
  \bibinfo{journal}{Stochastic Processes and their Applications}
  \textbf{\bibinfo{volume}{122}}, \bibinfo{pages}{2521} (\bibinfo{year}{2012}),
  ISSN \bibinfo{issn}{0304-4149},
  \urlprefix\url{https://www.sciencedirect.com/science/article/pii/S0304414912000671}.

\bibitem[{\citenamefont{Ditlevsen and Samson}(2019)}]{hypoellyptic-DitSamson}
\bibinfo{author}{\bibfnamefont{S.}~\bibnamefont{Ditlevsen}} \bibnamefont{and}
  \bibinfo{author}{\bibfnamefont{A.}~\bibnamefont{Samson}},
  \bibinfo{journal}{Journal of the Royal Statistical Society: Series B
  (Statistical Methodology)} \textbf{\bibinfo{volume}{81}},
  \bibinfo{pages}{361} (\bibinfo{year}{2019}),
  \eprint{https://rss.onlinelibrary.wiley.com/doi/pdf/10.1111/rssb.12307},
  \urlprefix\url{https://rss.onlinelibrary.wiley.com/doi/abs/10.1111/rssb.12307}.

\bibitem[{\citenamefont{Clairon and Samson}(2020)}]{clairon-samson}
\bibinfo{author}{\bibfnamefont{Q.}~\bibnamefont{Clairon}} \bibnamefont{and}
  \bibinfo{author}{\bibfnamefont{A.}~\bibnamefont{Samson}},
  \bibinfo{journal}{Statistical Inference for Stochastic Processes}
  \textbf{\bibinfo{volume}{23}}, \bibinfo{pages}{105} (\bibinfo{year}{2020}).

\bibitem[{\citenamefont{Pokern et~al.}(2009)\citenamefont{Pokern, Stuart, and
  Wiberg}}]{pokern}
\bibinfo{author}{\bibfnamefont{Y.}~\bibnamefont{Pokern}},
  \bibinfo{author}{\bibfnamefont{A.~M.} \bibnamefont{Stuart}},
  \bibnamefont{and} \bibinfo{author}{\bibfnamefont{P.}~\bibnamefont{Wiberg}},
  \bibinfo{journal}{Journal of the Royal Statistical Society: Series B
  (Statistical Methodology)} \textbf{\bibinfo{volume}{71}}, \bibinfo{pages}{49}
  (\bibinfo{year}{2009}),
  \eprint{https://rss.onlinelibrary.wiley.com/doi/pdf/10.1111/j.1467-9868.2008.00689.x},
  \urlprefix\url{https://rss.onlinelibrary.wiley.com/doi/abs/10.1111/j.1467-9868.2008.00689.x}.

\bibitem[{\citenamefont{Elerian et~al.}(2001)\citenamefont{Elerian, Chib, and
  Shephard}}]{Elerian}
\bibinfo{author}{\bibfnamefont{O.}~\bibnamefont{Elerian}},
  \bibinfo{author}{\bibfnamefont{S.}~\bibnamefont{Chib}}, \bibnamefont{and}
  \bibinfo{author}{\bibfnamefont{N.}~\bibnamefont{Shephard}},
  \bibinfo{journal}{Econometrica} \textbf{\bibinfo{volume}{69}},
  \bibinfo{pages}{959} (\bibinfo{year}{2001}),
  \eprint{https://onlinelibrary.wiley.com/doi/pdf/10.1111/1468-0262.00226},
  \urlprefix\url{https://onlinelibrary.wiley.com/doi/abs/10.1111/1468-0262.00226}.

\bibitem[{\citenamefont{Eraker}(2001)}]{Eraker}
\bibinfo{author}{\bibfnamefont{B.}~\bibnamefont{Eraker}},
  \bibinfo{journal}{Journal of Business \& Economic Statistics}
  \textbf{\bibinfo{volume}{19}}, \bibinfo{pages}{177} (\bibinfo{year}{2001}),
  \urlprefix\url{https://ideas.repec.org/a/bes/jnlbes/v19y2001i2p177-91.html}.

\bibitem[{\citenamefont{Zwanzig}(2001)}]{Zwanzig-book}
\bibinfo{author}{\bibfnamefont{R.}~\bibnamefont{Zwanzig}},
  \emph{\bibinfo{title}{Nonequilibrium statistical mechanics}}
  (\bibinfo{publisher}{Oxford University Press, USA}, \bibinfo{year}{2001}).

\bibitem[{\citenamefont{Miguel and Sancho}(1980)}]{Miguel1980}
\bibinfo{author}{\bibfnamefont{M.~S.} \bibnamefont{Miguel}} \bibnamefont{and}
  \bibinfo{author}{\bibfnamefont{J.~M.} \bibnamefont{Sancho}},
  \bibinfo{journal}{Journal of Statistical Physics}
  \textbf{\bibinfo{volume}{22}}, \bibinfo{pages}{605} (\bibinfo{year}{1980}),
  ISSN \bibinfo{issn}{1572-9613},
  \urlprefix\url{https://doi.org/10.1007/BF01011341}.

\bibitem[{\citenamefont{Kalman}(1960)}]{Kalman}
\bibinfo{author}{\bibfnamefont{R.~E.} \bibnamefont{Kalman}},
  \bibinfo{journal}{Journal of Basic Engineering}
  \textbf{\bibinfo{volume}{82}}, \bibinfo{pages}{35} (\bibinfo{year}{1960}),
  ISSN \bibinfo{issn}{0021-9223},
  \eprint{https://asmedigitalcollection.asme.org/fluidsengineering/article-pdf/82/1/35/5518977/35\_1.pdf},
  \urlprefix\url{https://doi.org/10.1115/1.3662552}.

\bibitem[{\citenamefont{Gillespie}(1996)}]{Gillespie}
\bibinfo{author}{\bibfnamefont{D.~T.} \bibnamefont{Gillespie}},
  \bibinfo{journal}{Phys. Rev. E} \textbf{\bibinfo{volume}{54}},
  \bibinfo{pages}{2084} (\bibinfo{year}{1996}),
  \urlprefix\url{https://link.aps.org/doi/10.1103/PhysRevE.54.2084}.

\bibitem[{\citenamefont{Brockwell and Davis}(2002)}]{Brockwell-Davis-book}
\bibinfo{author}{\bibfnamefont{P.}~\bibnamefont{Brockwell}} \bibnamefont{and}
  \bibinfo{author}{\bibfnamefont{R.}~\bibnamefont{Davis}},
  \emph{\bibinfo{title}{Introduction to Time Series and Forecasting}}
  (\bibinfo{publisher}{Springer, Berlin}, \bibinfo{year}{2002}).

\bibitem[{\citenamefont{Kadanoff}(1976)}]{Migdal-Kadanoff}
\bibinfo{author}{\bibfnamefont{L.~P.} \bibnamefont{Kadanoff}},
  \bibinfo{journal}{Annals of Physics} \textbf{\bibinfo{volume}{100}},
  \bibinfo{pages}{359} (\bibinfo{year}{1976}).

\bibitem[{\citenamefont{Parisi}(1988)}]{Parisi-book}
\bibinfo{author}{\bibfnamefont{G.}~\bibnamefont{Parisi}},
  \emph{\bibinfo{title}{{Statistical field theory}}}, Frontiers in Physics
  (\bibinfo{publisher}{Addison-Wesley}, \bibinfo{address}{Redwood City, CA},
  \bibinfo{year}{1988}), \urlprefix\url{https://cds.cern.ch/record/111935}.

\bibitem[{\citenamefont{Br\'ezin}(2010)}]{Brezin}
\bibinfo{author}{\bibfnamefont{E.}~\bibnamefont{Br\'ezin}},
  \emph{\bibinfo{title}{Introduction to Statistical Field Theory}}
  (\bibinfo{publisher}{Cambridge University Press}, \bibinfo{year}{2010}).

\bibitem[{\citenamefont{Skeel and Izaguirre}(2002)}]{LI}
\bibinfo{author}{\bibfnamefont{R.~D.} \bibnamefont{Skeel}} \bibnamefont{and}
  \bibinfo{author}{\bibfnamefont{J.~A.} \bibnamefont{Izaguirre}},
  \bibinfo{journal}{Molecular Physics} \textbf{\bibinfo{volume}{100}},
  \bibinfo{pages}{3885} (\bibinfo{year}{2002}),
  \eprint{https://doi.org/10.1080/0026897021000018321},
  \urlprefix\url{https://doi.org/10.1080/0026897021000018321}.

\bibitem[{\citenamefont{Gloter}(2001)}]{GloterOU}
\bibinfo{author}{\bibfnamefont{A.}~\bibnamefont{Gloter}},
  \bibinfo{journal}{Statistics} \textbf{\bibinfo{volume}{35}},
  \bibinfo{pages}{225} (\bibinfo{year}{2001}),
  \eprint{https://doi.org/10.1080/02331880108802733},
  \urlprefix\url{https://doi.org/10.1080/02331880108802733}.

\bibitem[{\citenamefont{Kubo}(1966)}]{Kubo}
\bibinfo{author}{\bibfnamefont{R.}~\bibnamefont{Kubo}},
  \bibinfo{journal}{Reports on Progress in Physics}
  \textbf{\bibinfo{volume}{29}}, \bibinfo{pages}{255} (\bibinfo{year}{1966}),
  \urlprefix\url{https://doi.org/10.1088/0034-4885/29/1/306}.

\bibitem[{\citenamefont{Marconi et~al.}(2008)\citenamefont{Marconi, Puglisi,
  Rondoni, and Vulpiani}}]{Vulpiani-response}
\bibinfo{author}{\bibfnamefont{U.~M.~B.} \bibnamefont{Marconi}},
  \bibinfo{author}{\bibfnamefont{A.}~\bibnamefont{Puglisi}},
  \bibinfo{author}{\bibfnamefont{L.}~\bibnamefont{Rondoni}}, \bibnamefont{and}
  \bibinfo{author}{\bibfnamefont{A.}~\bibnamefont{Vulpiani}},
  \bibinfo{journal}{Physics Reports} \textbf{\bibinfo{volume}{461}},
  \bibinfo{pages}{111} (\bibinfo{year}{2008}), ISSN \bibinfo{issn}{0370-1573},
  \urlprefix\url{https://www.sciencedirect.com/science/article/pii/S0370157308000768}.

\bibitem[{\citenamefont{Gradziuk et~al.}(2021)\citenamefont{Gradziuk,
  Torregrosa, and Broedersz}}]{Broedersz-Markov}
\bibinfo{author}{\bibfnamefont{G.}~\bibnamefont{Gradziuk}},
  \bibinfo{author}{\bibfnamefont{G.}~\bibnamefont{Torregrosa}},
  \bibnamefont{and} \bibinfo{author}{\bibfnamefont{C.~P.}
  \bibnamefont{Broedersz}}, \emph{\bibinfo{title}{Irreversibility in linear
  systems with colored noise}} (\bibinfo{year}{2021}), \eprint{2111.07359}.

\bibitem[{\citenamefont{Takens}(1981)}]{Takens}
\bibinfo{author}{\bibfnamefont{F.}~\bibnamefont{Takens}},
  \emph{\bibinfo{title}{Detecting strange attractors in turbulence}}
  (\bibinfo{publisher}{Springer-Verlag}, \bibinfo{year}{1981}), vol.
  \bibinfo{volume}{898}, pp. \bibinfo{pages}{366--381}.

\bibitem[{\citenamefont{Casdagli et~al.}(1991)\citenamefont{Casdagli, Eubank,
  Farmer, and Gibson}}]{Casdagli}
\bibinfo{author}{\bibfnamefont{M.}~\bibnamefont{Casdagli}},
  \bibinfo{author}{\bibfnamefont{S.}~\bibnamefont{Eubank}},
  \bibinfo{author}{\bibfnamefont{J.}~\bibnamefont{Farmer}}, \bibnamefont{and}
  \bibinfo{author}{\bibfnamefont{J.}~\bibnamefont{Gibson}},
  \bibinfo{journal}{Physica D: Nonlinear Phenomena}
  \textbf{\bibinfo{volume}{51}}, \bibinfo{pages}{52 } (\bibinfo{year}{1991}),
  ISSN \bibinfo{issn}{0167-2789},
  \urlprefix\url{http://www.sciencedirect.com/science/article/pii/016727899190222U}.

\bibitem[{\citenamefont{Stark}(1999)}]{Stark1}
\bibinfo{author}{\bibfnamefont{J.}~\bibnamefont{Stark}},
  \bibinfo{journal}{Journal of Nonlinear Science} \textbf{\bibinfo{volume}{9}},
  \bibinfo{pages}{255} (\bibinfo{year}{1999}).

\bibitem[{\citenamefont{Stark et~al.}(2003)\citenamefont{Stark, Broomhead,
  Davies, and Huke}}]{Stark2}
\bibinfo{author}{\bibfnamefont{J.}~\bibnamefont{Stark}},
  \bibinfo{author}{\bibfnamefont{D.~S.} \bibnamefont{Broomhead}},
  \bibinfo{author}{\bibfnamefont{M.~E.} \bibnamefont{Davies}},
  \bibnamefont{and} \bibinfo{author}{\bibfnamefont{J.}~\bibnamefont{Huke}},
  \bibinfo{journal}{Journal of Nonlinear Science}
  \textbf{\bibinfo{volume}{13}}, \bibinfo{pages}{519} (\bibinfo{year}{2003}).

\bibitem[{\citenamefont{Onsager and Machlup}(1953)}]{OnsagerMachlup53}
\bibinfo{author}{\bibfnamefont{L.}~\bibnamefont{Onsager}} \bibnamefont{and}
  \bibinfo{author}{\bibfnamefont{S.}~\bibnamefont{Machlup}},
  \bibinfo{journal}{Phys. Rev.} \textbf{\bibinfo{volume}{91}},
  \bibinfo{pages}{1505} (\bibinfo{year}{1953}),
  \urlprefix\url{https://link.aps.org/doi/10.1103/PhysRev.91.1505}.

\bibitem[{\citenamefont{Borra and Baldovin}(2021)}]{BorraBaldovin-chaos}
\bibinfo{author}{\bibfnamefont{F.}~\bibnamefont{Borra}} \bibnamefont{and}
  \bibinfo{author}{\bibfnamefont{M.}~\bibnamefont{Baldovin}},
  \bibinfo{journal}{Chaos: An Interdisciplinary Journal of Nonlinear Science}
  \textbf{\bibinfo{volume}{31}}, \bibinfo{pages}{023102}
  (\bibinfo{year}{2021}), \eprint{https://doi.org/10.1063/5.0036809},
  \urlprefix\url{https://doi.org/10.1063/5.0036809}.

\bibitem[{\citenamefont{Costa et~al.}(2021)\citenamefont{Costa, Ahamed, Jordan,
  and Stephens}}]{ACC}
\bibinfo{author}{\bibfnamefont{A.~C.} \bibnamefont{Costa}},
  \bibinfo{author}{\bibfnamefont{T.}~\bibnamefont{Ahamed}},
  \bibinfo{author}{\bibfnamefont{D.}~\bibnamefont{Jordan}}, \bibnamefont{and}
  \bibinfo{author}{\bibfnamefont{G.}~\bibnamefont{Stephens}},
  \emph{\bibinfo{title}{Maximally predictive ensemble dynamics from data}}
  (\bibinfo{year}{2021}), \eprint{2105.12811}.

\bibitem[{\citenamefont{Gardiner}(2009)}]{Gardiner}
\bibinfo{author}{\bibfnamefont{C.}~\bibnamefont{Gardiner}},
  \emph{\bibinfo{title}{Stochastic Methods}}, vol.~\bibinfo{volume}{13} of
  \emph{\bibinfo{series}{0172-7389}} (\bibinfo{publisher}{Springer-Verlag
  Berlin Heidelberg}, \bibinfo{year}{2009}).

\bibitem[{\citenamefont{Lee and Pressé}(2012)}]{LeePresse}
\bibinfo{author}{\bibfnamefont{J.}~\bibnamefont{Lee}} \bibnamefont{and}
  \bibinfo{author}{\bibfnamefont{S.}~\bibnamefont{Pressé}},
  \bibinfo{journal}{The Journal of Chemical Physics}
  \textbf{\bibinfo{volume}{137}}, \bibinfo{pages}{074103}
  (\bibinfo{year}{2012}), \eprint{https://doi.org/10.1063/1.4743955},
  \urlprefix\url{https://doi.org/10.1063/1.4743955}.

\end{thebibliography}
